ActivityDiff: A diffusion model with Positive and Negative Activity Guidance for De Novo Drug Design

Renyi Zhou[1,†], Huimin Zhu[1,†], Jing Tang[2] and Min Li[1,*]

1 School of Computer Science and Engineering, Central South University, Changsha 410083, China

[2] Faculty of Medicine, University of Helsinki, Helsinki, 00290, Finland

† These two authors contribute equally to the work.

* Corresponding author, limin@mail.csu.edu.cn

# Abstract

Achieving precise control over a molecule's biological activity—encompassing targeted activation/inhibition, cooperative multi-target modulation, and off-target toxicity mitigation—remains a critical challenge in de novo drug design. However, existing generative methods primarily focus on producing molecules with a single desired activity, lacking integrated mechanisms for the simultaneous management of multiple intended and unintended molecular interactions. Here, we propose ActivityDiff, a generative approach based on the classifier-guidance technique of diffusion models. It leverages separately trained drug-target classifiers for both positive and negative guidance, enabling the model to enhance desired activities while minimizing harmful off-target effects. Experimental results show that ActivityDiff effectively handles essential drug design tasks, including single-/dual-target generation, fragment-constrained dual-target design, selective generation to enhance target specificity, and reduction of off-target effects. These results demonstrate the effectiveness of classifier-guided diffusion in balancing efficacy and safety in molecular design. Overall, our work introduces a novel paradigm for achieving integrated control over molecular activity, and provides ActivityDiff as a versatile and extensible framework.

# Introduction

The biological behavior of a drug plays a critical role in determining its therapeutic efficacy and safety. Rational drug design aims to achieve multiple objectives: maximizing on-target efficacy, minimizing toxic off-target effects, and, in many cases, optimizing interactions with multiple therapeutic targets[1-3]. The ability to modulate multiple disease-relevant targets is often associated with improved therapeutic outcomes, particularly in complex diseases such as cancer or neurodegenerative disorders[4]. To meet these demands, molecular design strategies are expected to flexibly optimize biological activity across multiple targets while controlling off-target risks.

Deep learning models show significant potential in drug-target interaction prediction[5] and de novo drug design[6-10], leading to improving efficiency and reduced development costs. Some methods are primarily used to generate molecules that meet specific chemical properties[11, 12]. Other generative methods have demonstrated great potential in navigating the complex drug-like chemical space, thereby improving the efficiency of discovering molecules with therapeutic activity[13, 14]. However, generative methods typically focus on unidirectional optimization of activities. For example, Pocket2mol[6] utilizes the pocket structure as condition to generate molecules that binds the pocket. KGDiff[15] uses a network that predicts Vina score to guide the generation process and generates molecules with improved docking score. DeepDTAGen[16] is a multi-task deep learning model designed to simultaneously predict drug-target binding affinities and generate new drug candidates. These methods are typically confined to single-target drug design and cannot generate compounds under multi-target constraints.

A few approaches have been proposed to address the multi-target drug design challenge. A straightforward strategy involves connecting fragments known to be active against different targets, such as DeLinker[9], SyntaLinker[17], DEVELOP[18], and DiffLinker[19]. However, fragment linking approaches often suffer from the drawback of producing molecules with excessively large molecular weights. Alternatively, other de novo methods have been developed to tackle multi-target drug design from different perspectives. AIxFuse[20] learns pharmacophore fusion patterns to satisfy dual-target structural constraints in molecular docking simulations. POLYGON[21] is a multi-pharmacology method based on generative reinforcement learning that predicts compounds capable of inhibiting multiple proteins while maintaining high drug similarity and synthetic feasibility by embedding chemical space and iteratively sampling to generate new molecular structures. DRAGONFLY[22] integrates protein–protein interactions, drug–target interactions, and drug–drug relationships to construct an interaction network, and employs a graph neural network to generate molecules conditioned on this network.

In addition to optimizing activity against intended targets, it is also crucial to suppress unintended interactions with other targets, thereby improving binding selectivity and reducing the risk of off-target effects and downstream toxicity. Analyses of clinical failures from 2010–2017 reveal that while 40%–50% of candidates fail due to insufficient efficacy, 30% are abandoned for safety issues, among which unanticipated off-target interactions represent an important contributing factor[23]. For example, torcetrapib, a CETP inhibitor, was terminated in late-stage clinical trials due to off-target activation of the mineralocorticoid receptor, which led to increased cardiovascular events[24, 25]. These researches highlight the urgent need for methods that not only improve target affinity but also explicitly suppress harmful off-target effects.

Despite recent advances in molecular generative models, most methods focus on generating molecules that meet the desired positive activity, but fail to adapt to both single-target and multi-target scenarios, accounting for off-target effects.

In this work, we propose ActivityDiff, a diffusion-based molecular generative framework aimed at enabling selective, off-target minimized, and multi-target molecular generation. It uses pre-trained drug–target interaction classifiers to guide each step of the reverse diffusion process, providing fine-grained control over molecular generation toward predefined pharmacological objectives. Through positive prompting, the model is encouraged to generate molecules with high affinity for desired targets. Negative prompting is introduced to suppress affinities toward off-target proteins and reduce potential toxicity. Moreover, ActivityDiff supports multi-targets generation by allowing simultaneous application of multiple prompts, enabling both multi-target generation and selective generation between closely related targets. We evaluate the model's controllability and generalizability through diverse generation experiments, including single-target design, dual-target de novo generation, fragment-constrained dual-target generation, selective generation to improve target specificity, and generation minimizing off-target activity. In all cases, the model consistently produces molecules that align with the desired activity profiles specified by the classifiers, demonstrating strong adaptability to different design objectives. Together, these capabilities establish ActivityDiff as a flexible and extensible platform for activity-aware molecular generation, with broad potential to support rational drug discovery under complex pharmacological constraints.

# Results

## ActivityDiff overview

ActivityDiff extends the original classifier-guidance approach by introducing the negative guidance, where separately trained classifiers are used during the denoising process to push the intermediate results towards target properties. Classifier-guidance allows partial decoupling between generation and guidance, where incorporating new conditioning does not require retraining the generative model. In our study, we exploit classifier-guidance for active-like molecular generation by incorporating not only single positive guidance (e.g., promoting single target activity) but also negative guidance (e.g., suppressing off-target effects) and composite multi-object guidance (e.g., promoting multi-target activities, prompting selective target activity, etc.).

Figure 1 shows an illustration of ActivityDiff. The conceptual model of ActivityDiff can be seen in Figure 1a, the pharmacokinetic properties of a small molecule are determined by its structure, which mainly consists of atoms and bonds. Like previous studies[26], ActivityDiff corrupt the molecule represented by a sequence of discrete features independently during the diffusion forward process. Specifically, we treat molecules as complete graphs where each edge and node can be represented by a one-hot vector (Figure 1b), such that a molecule is transformed into a sequence of discrete feature vectors. We use discrete denoising diffusion models[27] to better capture the discrete nature of molecular space and enable fine-grid control over the generation process. During the denoising process, starting from a combination of randomly sampled atoms and bonds characterized by one-hot feature vectors, ActivityDiff progressively denoises them to yield a valid molecule at the final step. Separately trained classifiers which predicts whether a noisy input molecule has certain properties are used to modify the molecule in each denoising step (Figure 1c). Model structure, the altered guided denoising process, and other training details are further explained in the Method section.

ActivityDiff adapts to the diverse requirements of drug design tasks through different guidance setups to lead the generation process for multiple objectives. Figure 1d showcases several applications and their setups including single-target molecular generation, fragment-constrained generation, multi-target generation, specificity enhancement among homologous proteins, and avoiding off-target effects while designing bioactive molecules.

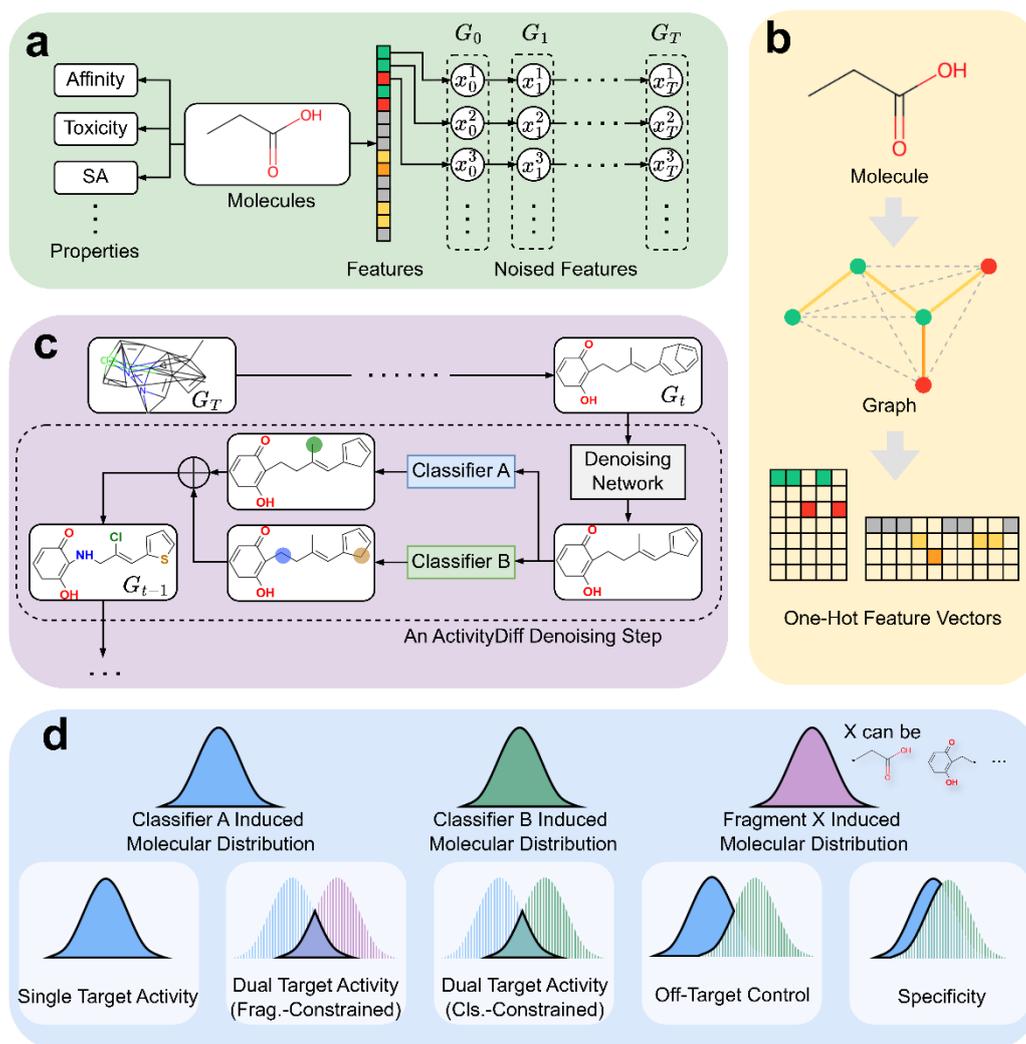

Figure 1. An illustration of ActivityDiff. **(a)** The dependency graphical model. Molecular attributes are determined by the molecular structure and the molecular structure can be decomposed as the collections of feature vectors. Each feature vectors are independently corrupted until reach the final state. **(b)** A molecule is considered as a complete graph and can be represented as the collection of node feature vectors and edge feature vectors. **(c)** An illustration of the denoising process. (d) A few downstream applications of ActivityDiff. The denoising process begins with a corrupted molecule sampled from noise. At each denoising step, the denoising network generates an intermediate denoised molecule. Next, the classifiers take the intermediate denoised molecule to calculate gradients based on the difference of the predicted value and the desired value of targeted attributes. The gradients are then used to guide the initial denoising step. See the method section for the actual computation process, which is a bit more complicated than the illustration. **(d)** Guidance setups for different drug design tasks.

## Performance of ActivityDiff in unconditional generation experiments

We benchmarked the performance of ActivityDiff in unconditional molecular generation setting against established baselines, including SMILES-based methods such as VAE[28], ORGAN[29], SMILES LSTM[30], Syntalinker[17], and PGMG[31], and graph-based methods such as Pocket2Mol[32], DEVELOP[18], LS-MolGen[33], and REINVENT2.0[34]. The results of Pocket2Mol, DEVELOP[18], LS-MolGen[33], and REINVENT2.0[34] are from the BDMA-MoGE benchmark. As shown in Table 1, ActivityDiff performs the best in terms of the ratio of available molecules, while achieving comparable levels of novelty, validity and uniqueness to other top models, such as Syntalinker, REINVENT2.0, PGMG, and SMILES LSTM. The superior unconditional generation performance indicates that ActivityDiff has a strong ability to generate novel molecules.

Table 1. Performance of ActivityDiff and other models in unconditional generation.

| Methods | Validity | Uniqueness | Novelty | Ratio of Available Molecules |
|---|---|---|---|---|
| ORGAN | 0.379 | 0.841 | 0.687 | 0.219 |
| VAE | 0.870 | 0.999 | 0.974 | 0.847 |
| SMILES LSTM | 0.959 | 1.000 | 0.912 | 0.875 |
| Syntalinker | 1.000 | 0.880 | 0.903 | 0.795 |
| PGMG | 0.948 | 0.999 | 1.000 | 0.948 |
| Pocket2Mol | 1.000 | 0.349 | 0.997 | 0.348 |
| DEVELOP | 0.856 | 0.334 | 0.997 | 0.286 |
| LS-MolGen | 0.632 | 1.000 | 0.761 | 0.481 |
| REINVENT2.0 | 0.963 | 0.999 | 0.999 | 0.961 |
| ActivityDiff | 0.978 | 0.999 | 0.999 | 0.975 |

The results of VAE[28], ORGAN[29], SMILES LSTM[30] are taken from GuacaMol[35].

## Generating active-like molecules under classifier guidance

We then evaluate the performance of these classifiers in guiding the generation of molecules, assessing the effect of positive and negative guidance. Positive guidance is intended to generate molecules predicted to be active against the target, whereas negative guidance aims to generate inactive compounds against the given target. We evaluate ActivityDiff on eight biological targets, as listed in Table 2, with detailed target information provided in Table S1 of the Supplementary Information. For each target, we generate 10000 molecules under both guidance modes and establish three control groups: (1) 10000 molecules randomly sampled from the generator's training set (GEOM[36]); (2) 10000 molecules randomly sampled from the classifier's training set

(BindingDB[37, 38]); and (3) 2000 molecules generated without guidance using ActivityDiff's base diffusion model. Classifier performance across different targets is provided in Table S2 of the Supplementary Information. All generated and control samples were evaluated using the trained classification models to predict their activity probabilities. The results are shown in Figure 2.

The results demonstrate that molecules generated under positive guidance exhibited a significantly higher proportion of predicted activity values exceeding 0.8 (Figure 2, red) compared to the control groups. Specifically, the positive guidance group achieved a significantly higher proportion of samples in the high-activity region ($Y \geq 0.5$), reaching 78.8% ± 16.3% (n = 8). In contrast, the GEOM dataset yielded 13.8% ± 15.8% (Figure 2, orange), the BindingDB dataset 21.7% ± 15.0% (Figure 2, purple), and the unconditional generation group only 8.1% ± 8.0% (Figure 2, green). These results demonstrate that classifier-guided generation effectively biases molecular output toward the desired activity space. To further evaluate the bidirectional controllability of ActivityDiff, we also examined the effect of negative guidance. The average prediction score of molecules generated under negative guidance is 0.04 ± 0.09 which is significantly lower ($p<0.001$) than the scores of molecules from BindingDB (0.246 ± 0.304), GEOM (0.172 ± 0.249), unconditionally generated molecules (0.126 ± 0.208), and those generated under positive guidance (0.679 ± 0.279). Furthermore, 80.4% of the molecules generated with negative guidance exhibit classifier scores below 0.1, in stark contrast to only 9.7% of the molecules generated with positive guidance falling into the same range. Together, these results demonstrate that ActivityDiff enables bidirectional control over molecular generation.

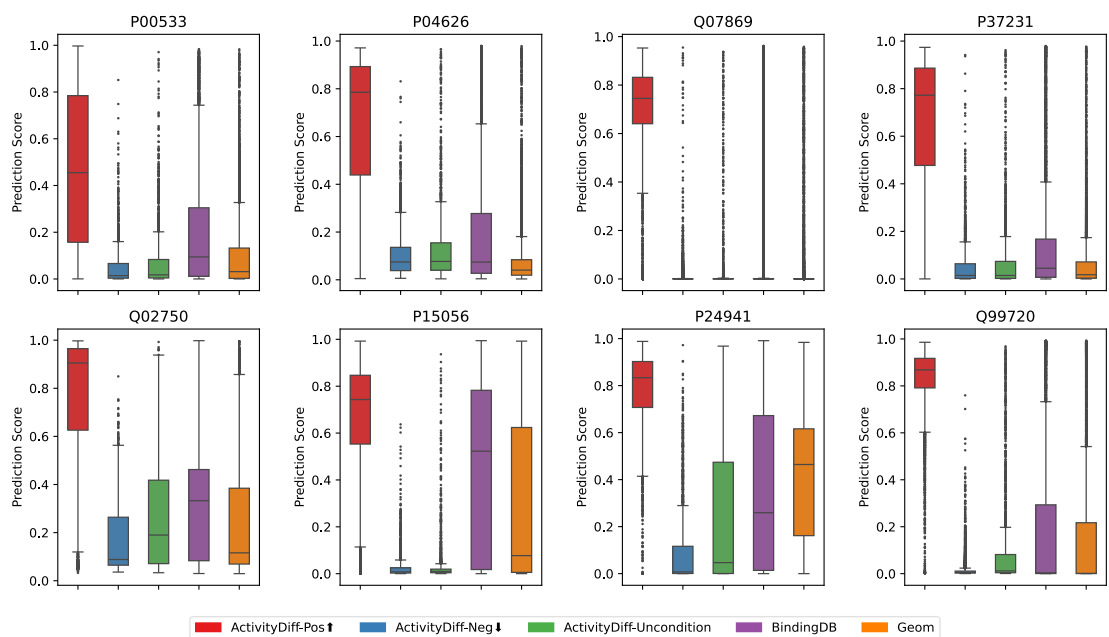

**Figure 2**. Boxplots showing the prediction scores based on the trained classifiers for molecules generated across different targets. Red represents for positive guidance, blue

for negative guidance, green for unconditioned generation, and purple for BindingDB dataset, orange for Geom dataset.

Though ActivityDiff is ligand-based and we are aware of the known discrepancies between structure-based docking results and ligand-based predictions[39], we conducted docking experiments to offer an additional perspective on our results and confirm the reliability of the selected candidates. For each target, we generates 10000 molecules using ActivityDiff with single positive guidance, and use AutoDock Vina[40] to calculate the docking scores of those molecules. We calculate the 'Pass Rate' for the generated molecules, which represents the proportion of molecules with docking scores exceeding a threshold set by the 95th percentile of scores from experimentally active molecules. This threshold is chosen to reduce the influence of extreme values and the active molecules are taken from the BindingDB database. We also calculate the average docking scores of those active molecules and top 1000 molecules generated by ActivityDiff. As shown in Table 2, the results indicate that a large proportion of the molecules generated for different targets are identified as being within the range of active molecules in terms of docking scores.

**Table 2. Docking performance of molecules generated by ActivityDiff and experimental molecules.**

| Target | Experiment | Top1000 | Pass Rate |
|---|---|---|---|
| P00533 (1M17) | $-8.336 \pm 0.811$ | $-9.638 \pm 0.535$ | 0.761 |
| P04626 (3PP0) | $-9.467 \pm 1.864$ | $-11.051 \pm 0.439$ | 0.884 |
| Q99720 (5HK1) | $-9.426 \pm 1.250$ | $-10.908 \pm 0.546$ | 0.784 |
| P37231 (1I7I) | $-8.994 \pm 1.080$ | $-9.241 \pm 0.499$ | 0.714 |
| Q02750 (1S9J) | $-8.434 \pm 1.031$ | $-9.155 \pm 0.461$ | 0.780 |
| P15056 (1UWH) | $-10.225 \pm 1.358$ | $-9.789 \pm 0.505$ | 0.542 |
| P24941 (4KD1) | $-7.852 \pm 0.935$ | $-8.702 \pm 0.407$ | 0.870 |
| Q07869 (1K7L) | $-8.558 \pm 0.799$ | $-8.863 \pm 0.495$ | 0.523 |

## Generating dual-target compounds

Melanoma is often associated with mutations in the NRAS and BRAF genes within the MEK signaling pathway. Research has shown that the combination of BRAF/MEK inhibitors shows significant synergy for BRAF-mutant cancers[41]. In this case, we trained two classifiers to identify active and inactive molecules targeting MEK and BRAF. The classifiers were then used in the ActivityDiff generation process to guide molecular generation. The results using dual positive guidance can be found in Figure 3.

Figure 3a shows the distribution of prediction score by MEK classifier of molecules generated under single-target guidance and those by using two classifiers. Specifically, molecules generated via BRAF-specific guidance (red) were evaluated using the MEK classifier, yielding predicted activity values predominantly within the 0.2–0.6 range (median 0.417 ± 0.044), which is significantly lower than those in the MEK single-target guidance group (blue, median 0.905 ± 0.064; $p < 0.01$). Notably, under the dual-target guidance mode (green distribution), the generated molecules exhibit a MEK activity score distribution (median 0.903 ± 0.055) that is statistically indistinguishable from the MEK single-target guidance group (Mann-Whitney U, $p = 0.505$), indicating that the dual-guidance system effectively preserves MEK-targeted activity.

Figure 3b shows the distribution of BRAF prediction scores for molecules generated under single-target guidance and dual guidance, as evaluated by the BRAF activity prediction model. The predicted values are predominantly in the low-activity range (0–0.3, median 0.30), exhibiting a significant difference from the BRAF single-target guidance group (which primarily falls within the 0.6–0.9 range, with a median of 0.743; $p < 0.01$). Interestingly, when molecules are generated under BRAF/MEK dual-classifier joint guidance (Figure 3b, green), their BRAF activity score median reaches 0.754, exhibiting a high degree of overlap >90% with the BRAF single-target guidance group (median 0.743), as quantified by the Bhattacharyya coefficient[42]. Overall, above results demonstrate that ActivityDiff enables multi-target guidance.

Figure 3 (c, d) illustrates the binding sites of reference molecules from PDB and dual-target molecules generated using the joint BRAF-MEK classifier, designed for MEK (UniProt ID: Q02750, PDB: 1S9J) and BRAF (UniProt ID: P15056, PDB: 1UWH). The binding pose of generated molecules are acquired using a docking program. Structural analyses of the binding sites reveal that classifier-guided molecules exhibit high compatibility with the MEK binding pocket and form well-aligned interactions within the BRAF binding site. Further examination indicates that both protein pockets engage in π–π stacking interactions and hydrogen bonding with the generated ligands, thereby enabling effective accommodation of both targets. These molecules achieve classifier scores exceeding 0.8 for both targets, and their docking scores are comparable to those of known protein–ligand complexes. These findings highlight the model's capability to generate dual-target molecules. The dual-classifier guidance strategy dynamically balances the structural requirements of MEK and BRAF, demonstrating its feasibility in designing synergistic dual-target inhibitors.

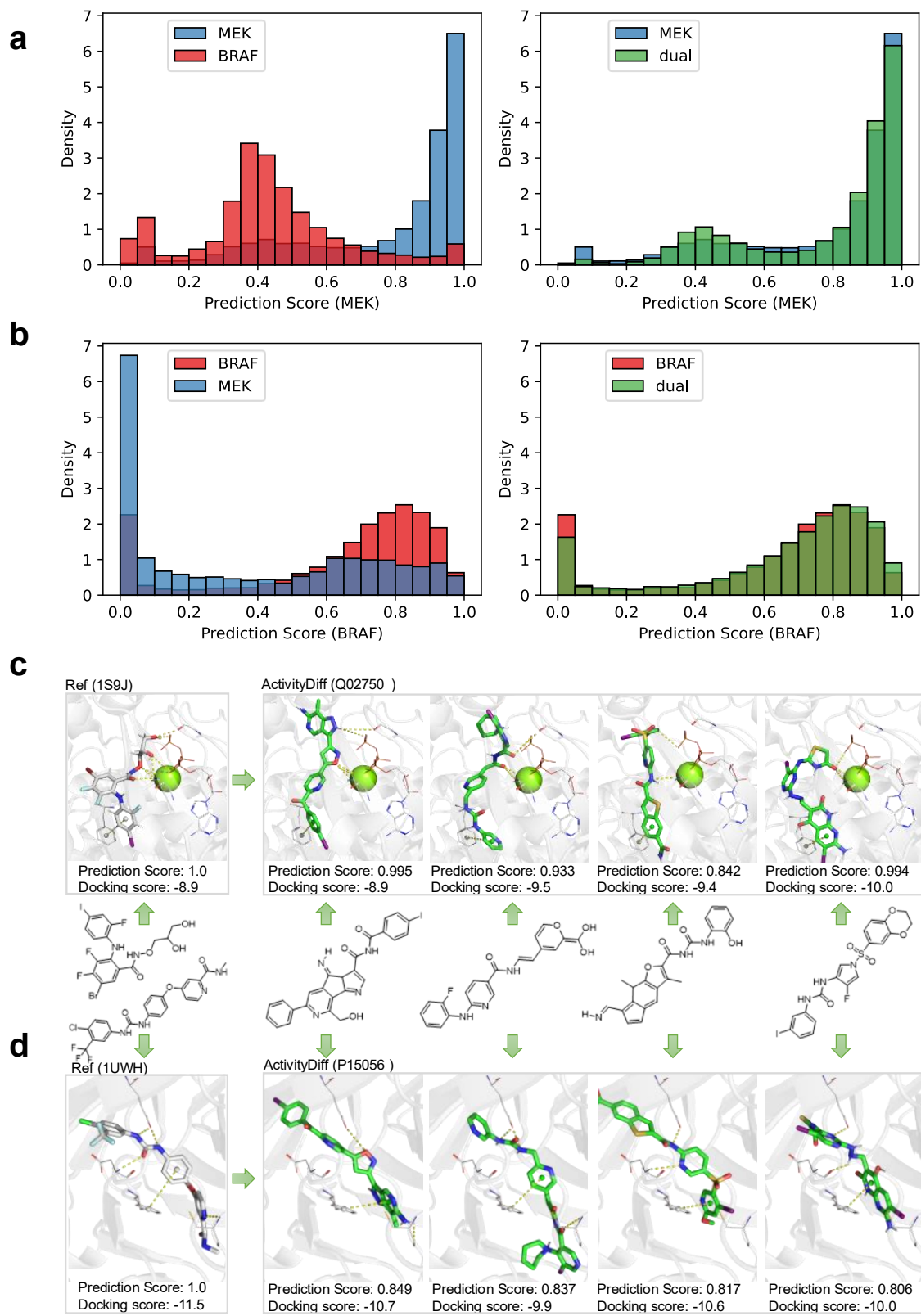

Figure 3. (a) MEK prediction scores for molecules generated under different guidance strategies: single target guidance (left); combined BRAF–MEK vs. MEK guidance (right). (b) BRAF prediction scores for molecules generated under different guidance

strategies: single target guidance (left); combined BRAF–MEK vs. MEK guidance (right). (c) MEK binding analysis of molecules generated using dual guidance: the prediction scores (MEK classifier), docking scores and binding conformations of reference (left, PDB: 1S9J) and generated molecules (right). (d) BRAF binding analysis of molecules generated using dual guidance: the prediction scores (BRAF classifier), docking scores and binding conformations of reference (left, PDB: 1UWH) and generated molecules (right). Yellow dashed lines indicate ligand–protein interactions.

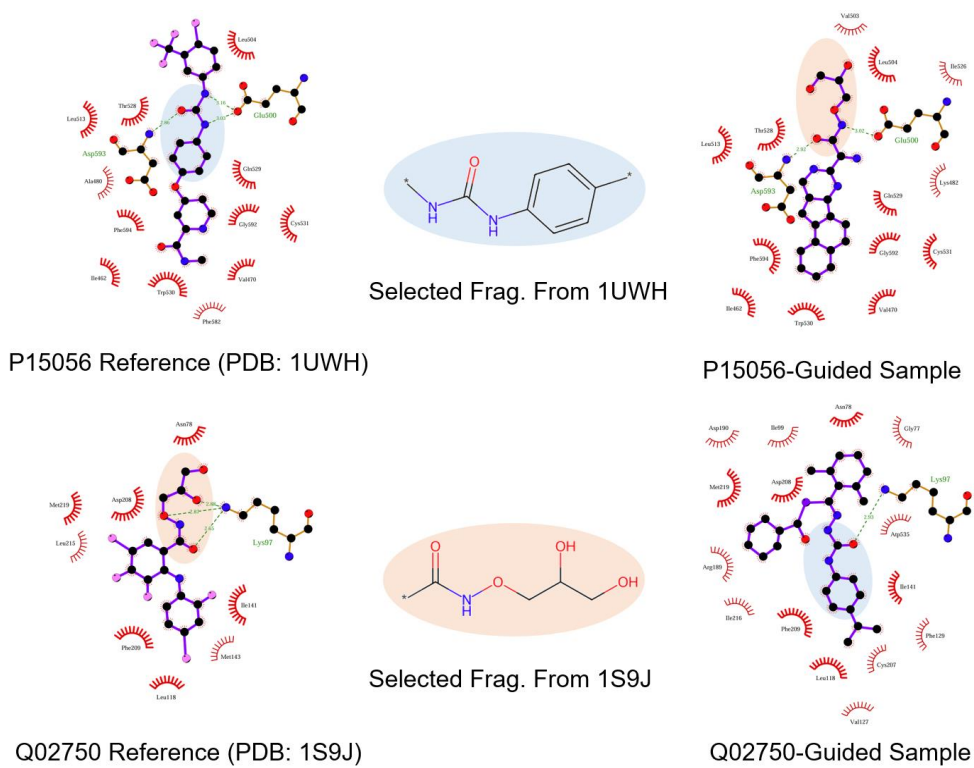

Figure 4. Dual-target generation results by guiding molecule generation toward one target while preserving the active fragment of another. The left panel shows the binding sites of the two targets extracted from the PDB complex structures. For each target, we selected one bioactive fragment from its ligand-bound complex structure and used it as a condition while guiding the generation process toward the other target, with the resulting molecules shown in the right panel. Additional molecules are provided in the Supporting Information (SI).

Additionally, since the design of base diffusion model enables generation using fixed fragments, we also evaluated ActivityDiff's dual-target drug design performance when guiding generation toward one target while preserving the active fragment of another (Figure 4). We did an extra dual-target design experiment by fixing the active fragment of one target and guiding the generation process towards another target. A demonstration is given in Figure 4. We manually selected a fragment from an active molecule in the PDB complex structure for each target and used that fragment as a condition combining with the classifier guidance towards another target to generate dual-targeting molecules. The generated molecules are able to preserve the given fragment and also yield good prediction score of another classifier.

## Generating specific inhibitors

HER2 (human epidermal growth factor receptor 2) and EGFR (epidermal growth factor receptor 1) are two highly homologous receptors with significant structural similarities. Although EGFR/HER2 combination inhibitors achieve favorable outcomes in the treatment of various cancers. These inhibitors are often associated with adverse effects such as diarrhea and rashes, which can severely impact the quality of life of patients. In contrast, HER2 specific-targeted drugs have shown distinct advantages in treating HER2-positive breast cancer[43]. By targeting HER2, these drugs can effectively inhibit tumor cell proliferation while avoiding EGFR-related side effects. In this study, the objective is to generate molecules with high activity toward HER2 (UniProt ID: P04626) while minimizing off-target interactions with EGFR (UniProt ID: P00533). To this end, two classifiers are trained to recognize compounds with activity against EGFR and HER2, respectively. During the ActivityDiff generation process, HER2 selectivity is promoted using a positive guidance against HER2 and a negative guidance against EGFR.

Figure 5a presents the prediction score distributions of molecules generated under HER2 classifier guidance, as evaluated by the HER2 classifier (blue) and the EGFR classifier (red). Figure 5b illustrates the distributions of molecules generated under combined HER2-positive and EGFR-negative guidance, also evaluated using the HER2 classifier (red) and the EGFR classifier (blue). 22.7% of molecules generated under HER2 classifier guidance has an EGFR prediction score of over 0.5 (Figure 5a, red), suggesting potential dual affinity. However, this proportion declines significantly to 6.4% (Figure 5b, red) when using the specificity-guided approach, which proves the efficacy of the specificity-guided approach.

Notably, HER2 classifier scores remain consistently high regardless of the specificity adjustment. The proportion of molecules with HER2 classifier scores above 0.5 reaches 69.3% for HER2 classifier guidance and 62.2% for specificity-guidance. On the other hand, we do observe a small shift of the distribution to the left, which might be explained by the high homology of the two targets. Figure 5c showcases representative molecular structures generated under specific classifier guidance, along with their prediction scores from the HER2 and EGFR classifiers and their molecular docking scores. These molecules show high HER2 prediction score and low EGFR prediction score. These findings demonstrate that ActivityDiff effectively enhances target specificity while preserving high HER2 activity, thereby offering a framework for selective molecular generation.

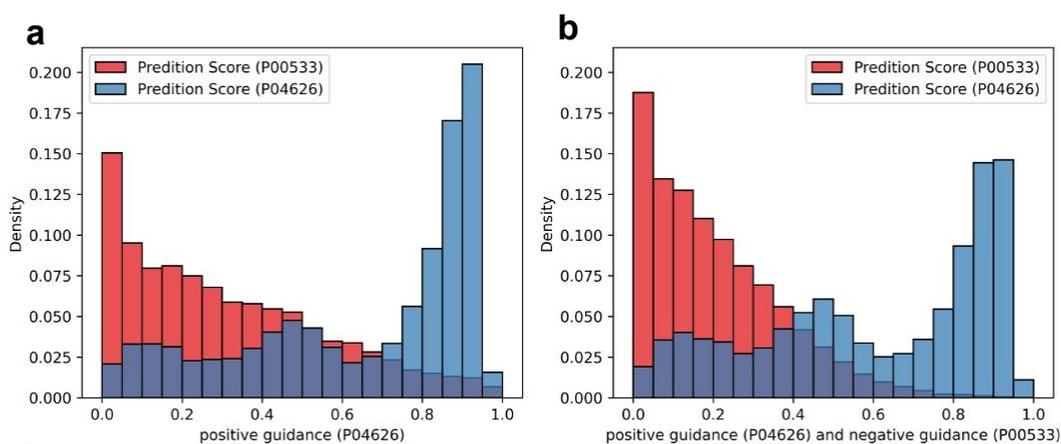

| Compound | Target | Prediction Score | Docking Score |
|---|---|---|---|
| | P04626(HER2) | 0.883 | -9.114 |
| | P00533(EGFR) | 0.293 | -6.849 |
| | P04626(HER2) | 0.858 | -9.002 |
| | P00533(EGFR) | 0.051 | -.6125 |
| | P04626(HER2) | 0.884 | -9.131 |
| | P00533(EGFR) | 0.042 | -6.954 |
| | P04626(HER2) | 0.907 | -9.289 |
| | P00533(EGFR) | 0.128 | -6.893 |

Figure 5. (a) shows the HER2 (UniProt ID: P04626, blue) prediction scores and the EGFR (UniProt ID: P00533, red) score of molecules generated by guiding with the HER2 classifier. (b) shows the HER2 (blue) prediction scores and the EGFR (red) score of molecules generated using combined positive guidance from the HER2 classifier and negative guidance from the EGFR classifier. (c) Exhibition of molecules generated via specificity generation.

## Reducing broad-spectrum off-target effects

Studies have shown that off-target interactions are a major factor contributing to adverse drug effects[23]. In preclinical drug development, off-target panels are commonly used to test preclinical molecules[44, 45]. Off-target panels screening against large panels of safety-relevant targets enables early identification of well-targeted compounds, reducing safety-related risks and improving drug success and viability. In this section, six targets are selected based on the sensitivity and hit rate analysis of the BioPrint1[46] dataset conducted by Peters et al.[47] which identifies targets that perform well in predicting compound promiscuous binding. These selected targets are highly representative across the entire off-target panel. For instance, 70% of compounds that bind to the 5-HT2B receptor are classified as promiscuous across the panel, whereas only 0.64% of compounds with no significant affinity (pIC50 < 3.5) identified promiscuous[48]. Therefore, the study considers these six targets to constitute a reduced off-target profile. A joint classifier is trained based on these six targets. A training sample is labeled as a positive if it exhibits high binding activity to any of the six targets, while samples showing binding affinity below 10,000 nM (Ki, Kd, IC50) to all six targets are labeled as negative.

The generation experiment aimed to produce target-selective molecules while not binding to the targets in the off-target panel. The generated molecules are then predicted using an external off-target predictor[44]. For comparison, experimentally validated active molecules are also assessed using the same predictor. As the off-target predictor covers multiple targets, only the highest prediction score for each molecule is retained. A classification threshold of 0.5 is applied. If the prediction score exceeds 0.5, the molecule is considered to interact with at least one target in the off-target panel.

As shown in Figure 6a, for all targets, the molecules generated with the off-target panel classifier guidance have a lower off-target ratio than the experimental active molecules. Figure 6b shows the percentage of molecules predicted to have off-target risk, based on an off-target prediction model, for both ActivityDiff-generated compounds and experimentally active molecules from BindingDB. "Improvement" reflects the reduction in off-target risk achieved by ActivityDiff compared to the experimental actives. As shown in the table, across the tested targets, the compounds generated by ActivityDiff consistently show a lower proportion of predicted off-target liabilities than the experimental molecules.

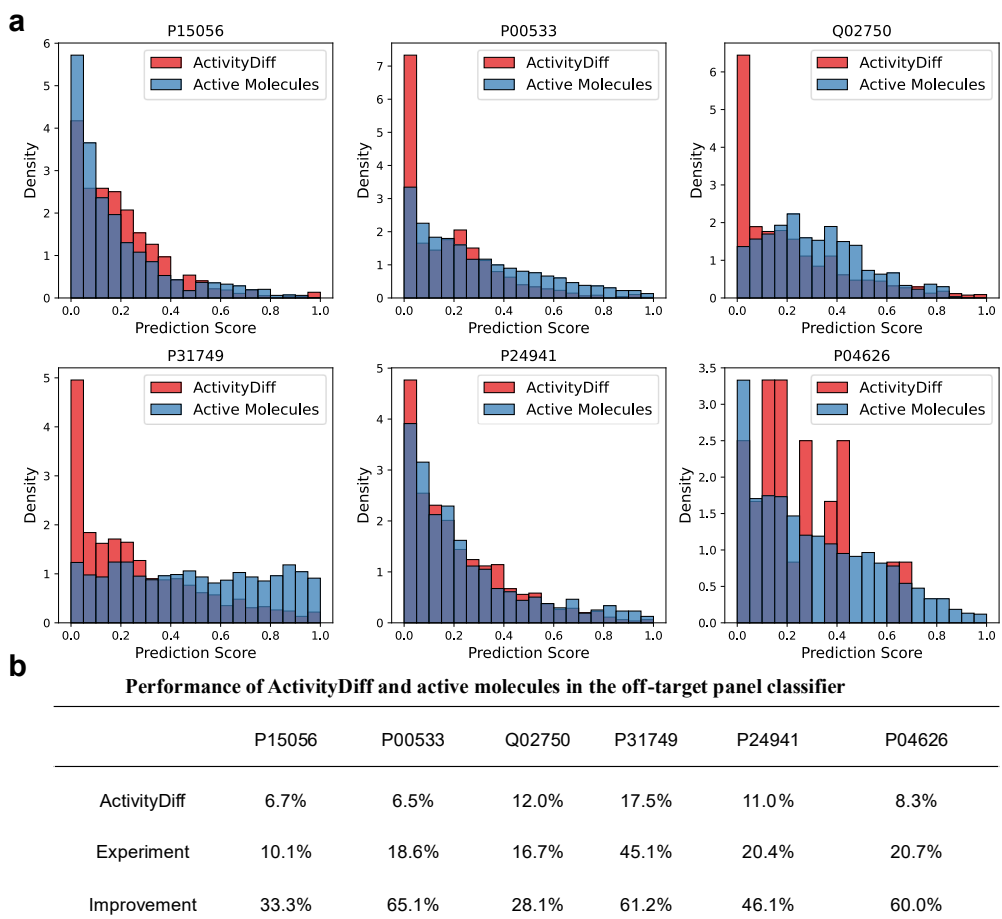

Figure 6. (a) Distribution of prediction scores in the off-target spectrum panel for ActivityDiff generated molecules (red) and experimental active molecules (blue). (b) Performance of ActivityDiff and active molecules in the off-target panel classifier

# Discussion

In this study, we propose ActivityDiff, a framework built upon discrete diffusion models with classifier-based guidance, designed to suppress off-target risks, enhance selectivity, and support multi-target generation. The separated trained classifier reduces the cost of leveraging the method in a new scenario. The classifiers are leveraged to provide positive guidance which improves desired properties and negative guidance which reduces undesired off-target effects. The simplicity of positive and negative classifier-guidance does not hinder the efficacy and flexibility of ActivityDiff.

The potential application of ActivityDiff spans diverse scenarios as demonstrated in our case studies. In single-target applications, it guides the generation of either active or inactive compounds, ensuring targeted optimization. When applied to multi-target drug design, ActivityDiff integrates multiple classifiers to facilitate the design of compounds with activity across different targets. Additionally, by incorporating negative guidance from undesired target classifiers, the approach enhances selectivity, minimizing off-target effects and toxicity. Furthermore, ActivityDiff's use of fragments allows for the design of multi-target active molecules, offering a refined and effective strategy for complex molecular designs that balance both activity and selectivity.

Despite these strengths, several challenges remain. Balancing affinities across multiple targets requires further investigation. Additionally, more effective utilization of inactive compound data, along with improved predictions of drug metabolism and toxicity, is essential for enhancing model reliability. Once a drug enters the body, it is exposed to a complex biological environment involving drug–drug interactions, metabolic transformations, and potential toxicity from metabolites. Addressing these factors is critical to ensuring the safety and efficacy of candidate drugs. In this context, integrating drug design into a systems-level biological network—taking into account drug–protein, protein–protein, and drug–drug interactions—represents a promising direction for future research and model refinement.

# Methods

## Discrete denoising diffusion

We represent a molecule with $N$ atoms as an undirected complete graph $G = \{V, E\}$, where $V = \{v, i = 1,2, \ldots, N\}$ is the set of vertices with each vertex characterized by its atom symbol and formal charge, and $E = \{e_j, j = 1,2, \ldots, \frac{N \times (N-1)}{2}\}$ is the set of edges with each edge characterized by its bond type (including non-bond) and the nodes it connects.

To enhance controllability during the molecular generation process, we adopt the discrete denoising diffusion framework for molecular generation. Discrete denoising diffusion models, like other diffusion models, consist of two key components: a forward process that add noise and a denoising reverse process. Following D3PM's[27] proposed framework, for a discrete random variable with K categories of the graph, $x \in \{1,2, \ldots, K\}$, the forward process $q$ progressively corrupts it and yields a sequence of corrupted variables $\{x_t\}_{t=0}^{T}$, where $x_0$ represents the original variable and $x_T$ aligns with the maximally corrupted state distribution $\pi$. Let $x_t$ be represented by a one-hot column vector, and the forward process at each step $t$ can be described by a categorical distribution $q(x_t \mid x_{t-1}) = Cat(x_t; P = Q_t x_{t-1})$, where $Q_t \in R^{K \times K}$ is the transition matrix.

We apply such a forward process independently to each categorical features in the graph G during the corruption process of a molecule, such that $q(x_t \mid G_{t-1}) = q(x_t \mid x_{t-1})$, where $x_t$ is a feature and $G_t$ represented the corrupted graph at time step $t$.

In the reverse process $p$, the aim is to reconstruct the original $G_0$ from its corrupted state $G_T$ by progressively denoising it, which is formulized as[49, 50]:

$$p_\theta(x_{t-1}|G_t) = \sum_{x_0} q(x_{t-1}|x_0, x_t) p_\theta(x_0|G_t),$$

where $q(x_{t-1}|x_0, x_t)$ is the forward process transition probabilities, and $p_\theta(x_0|G_t)$ is the estimated marginal distribution of $x_0$ given $G_t$. Note that the denoising neural network takes not only a single feature vector but the whole graph.

## Classifier-guidance with Discrete denoising diffusion

To guide the model in generating molecules with specific properties, we adopt the classifier-guidance approach[51]. Independently trained classifiers are leveraged to steer the generation process towards producing molecules with desired characteristics. The key advantage of this approach is its flexibility: the classifier is trained separately from the generator. This separation allows for greater adaptability in targeting specific molecular features without retraining the entire generation model which significantly reduces the cost of adapting ActivityDiff to a specific scenario.

As can be seen from the dependency graph (**Figure 1b**), the forward diffusion process remains unchanged under the conditional setting. During the reverse process, the objective is modified to incorporate the condition $y$, where y represents the desired molecular feature. The overall conditional objective becomes $p(G_{t-1}|G_t, y)$ and for each variable $x$, the objective be rewritten using Bayes' rule and the dependence relations as:

$$p(x_{t-1}|x_t, y) \propto p_\theta(x_{t-1}|x_t) p_\phi(y|G_{t-1}), \qquad (1)$$

where $p_\theta(x_{t-1}|x_t)$ is the same with unconditional generation, and $p_\phi(y \mid G_{t-1})$ is a separately trained classifier that predicts the likelihood of the condition $y$ being satisfied given the corrupted molecular graph $G_{t-1}$ at time step $t - 1$. $p_\phi(y|G_{t-1})$ is then approximated by performing a first-order Taylor expansion around a reference graph $G^*_{t-1}$:

$$p(x_{t-1}|x_t, y) \propto p_\theta(x_{t-1}|x_t) \cdot p_\phi(y|G_{t-1})$$

$$\propto p_\theta(x_{t-1}|x_t) \cdot [p_\phi(y|G^*_{t-1}) + (G_{t-1} - G^*_{t-1})$$

$$\cdot \nabla_{G_{t-1}} p_\phi(y|G_{t-1})|_{G_{t-1}=G^*_{t-1}}]$$

$$\propto p_\theta(x_{t-1}|x_t) \cdot (x_{t-1} - x^*_{t-1}) \cdot \nabla_{G_{t-1}} p_\phi(y|G_{t-1})|_{G_{t-1}=G^*_{t-1}} \qquad (2)$$

Another way to decompose $p(x_{t-1}|x_t, y)$ is

$$p(x_{t-1}|x_t, y) = \sum_{x_0} q(x_{t-1}, x_t|x_0, y) p_\theta(G_0|G_t, y) \qquad (3)$$

Similarly, we can expand $p_\theta(G_0|G_t, y)$. In our model, both decomposing methods are used with same weights. By incorporating these two approximations into the reverse process, the model is guided to prioritize molecular graphs that satisfy the condition y.

## Model architecture

We extend the denoising network proposed by[52] to incorporate variable-level noise embedding into the denoising process. The denoising network takes a corrupted graphs as input and outputs its prediction of the uncorrupted graph. The network is composed of several attention blocks where node features are updated using FiLM[53] layers with respect to global features and edges features while edge features are updated using the attention score between node features. The initial noise levels for node and edge features are first transformed into dense conditioning vectors, denoted as $c_v$ and $c_e$ respectively. These vectors are generated through dedicated multi-layer perceptrons (MLPs) from their corresponding noise levels. Within each of the L attention layers, features $F^L = \{f_v^L, f_e^L, f_g^L\}$ undergo an adaptive normalization step, which utilizes the conditioning vectors $C = (c_v, c_e)$ to incorporate the noise levels. Following normalization, the multi-head attention block processes the features. The resulting updates from the attention blocks are then integrated into the main feature streams via gated residual connections. For a given feature type $X \in (v, e)$, the update rule is generalized as:

$$f_X^{L+1} = f_X^L + G_X(c_X) \odot Attention_X(Norm_X(f_X^L, c_X)) \qquad (4)$$

Here, $Norm_X(\cdot)$ denotes the adaptive normalization layer, $Attention_X(\cdot)$ represents the output of the multi-head attention block for feature X, and $G\_X(\cdot)$ is a gate layer. Global features $f_g$ are updated within the attention block itself. In total, the network contains 12 layers of multi-head attention blocks with 4 heads. The hidden dimension of node, edge, and global features are set to 128, 64, and 128 respectively.

We choose the MPNN from dgllife[54] as the structure of the classifiers. It contains a graph neural network encoder consisting of several layers of MPNN variants proposed by Lu[55], a Set2Set pooling[56] layer and a MLP to produce the final prediction. This model leverages atomic features and formal charges as node attributes, and bond types as edge features. Molecular graphs are encoded via graph convolutional layers that aggregate information from neighboring nodes through a message-passing mechanism, enabling the model to learn expressive node-level representations. These representations are subsequently aggregated and passed through linear layers for final classification.

## Training Details

The denoising network is trained using GEOM[36]. Given the imbalanced distribution of atom types and other features in the dataset and real world, in the experiment, we only consider 30 atom types (H, B, C, N, O, F, Al, Si, P, S, Cl, As, Br, I, Hg, Bi, Se, Fe, Pt, V, Rh, Co, Ru, Mg). Molecules with other types of atoms are excluded from training. We use kekulized molecules, so there are four edge types (none, single_bond, double_bond, triple_bond) to be considered. In total, the denoising network contains 6.00 million parameters and each classifier 0.68 million parameters.

The classifiers are trained using drug-target interaction data taken from BindingDB. For each target, experimental compounds were collected from the BindingDB database and split into training, validation, and test sets in a ratio of 8:1:1. Labels were assigned based on compound activity: compounds with activity values less than 1000 nM were labeled as active (label 1), and those with values greater than or equal to 10,000 nM were labeled as inactive (label 0). Compounds with intermediate activity values were mapped to the [1, 0] interval by taking the natural logarithm of their activity values. To enhance the model's generalization ability, negative sampling was applied to the training data. Specifically, for each target, we randomly sampled inactive compounds with a structural similarity below 0.6 to the active compounds. This ensures the selected negative samples are structurally dissimilar, reducing label noise and improving discriminative learning. The final ratio of active to inactive compounds in the training set was maintained at 1:10.

To simulate intermediate states during the diffusion process, ActivityDiff injects random diffusion noise at step t into the input molecule. The pre-trained classifier then evaluates the noisy molecules at each diffusion step, providing guidance for the generation trajectory. To account for the varying informativeness of molecular features across noise levels, the classifier incorporates signal-to-noise ratio (SNR) weights to reweight the loss. Specifically, the loss function is a weighted binary cross-entropy, where the weight for each sample is determined by its corresponding SNR. The loss is defined as follows:

$$loss = -\frac{1}{N}\sum_{i=1}^{N}\left[y_i * \log(\sigma(z_i)) + (1-y_i) * \log(1-\sigma(z_i))\right] * snr\_weight_i \quad (5)$$

Here, $z_i$ denotes the model's output logits, $y_i$ represents the ground-truth label, and $\sigma(z_i)$ is the sigmoid function applied to the logits. $snr\_weight_i$ is the signal-to-noise ratio (SNR) weight for the $i-th$ sample, provided by the diffusion model. $N$ is the total number of samples.

During training, the Adam optimizer is employed with a learning rate of 0.0003 and no weight decay to update the model parameters. The maximum number of training epochs is set to 1000. To prevent overfitting, an early stopping strategy is applied with a patience value of 30. Training is terminated early if the performance on the validation set, measured by the AUC score, does not improve over a number of consecutive epochs.

# Data availability

Source data are provided with this paper.

# Code availability

Source code is provided in github.com/e-yi/ActivityDiff.

**Acknowledgements**

This work is financially supported by the National Natural Science Foundation of China under Grants (No. 62225209 to M.L.), Hunan Provincial Natural Science Foundation of China (2025JJ30025 to M.L.), European Research Council (No. 716063 to J.T.), and the High Performance Computing Center of Central South University.

**Author Contributions**

M.L. supervised the research and provided the experimental platform. H.Z. proposed the initial idea. R.Z designed the framework and developed the generation

model, H.Z. trained the classification model. H.Z and R.Z performed the experiments. M.L, H.Z, R.Z, J.T. analyzed the results and wrote the manuscript.

**Competing Interests Statement**

The authors declare no competing interests.